**Title**

deepNoC: A deep learning system to assign the number of contributors to a short tandem repeat DNA profile

**Author**

Duncan Taylor[1,2] (Duncan.Taylor@sa.gov.au), Melissa A. Humphries[3] (melissa.humphries@adelaide.edu.au)

1. Forensic Science SA, GPO Box 2790, Adelaide, SA 5001, Australia
2. College of Science & Engineering, Flinders University, GPO Box 2100, Adelaide, SA 5001, Australia
3. School of Computer and Mathematical Sciences, University of Adelaide, North Terrace Campus, Adelaide, SA, Australia 5005

**Corresponding author**

Duncan A. Taylor, PhD
Forensic Science SA
GPO Box 2790
Adelaide
South Australia 5001
Phone: +61-8 8226 7700
Fax: +61-8 8226 7777
Email: Duncan.Taylor@sa.gov.au

**Highlights**

- An analysis pipeline is developed to simulate pre-labelled training DNA profile data
- The profiles are used to train a neural network to assign the number of contributors
- The neural network achieves a high degree of accuracy in assignments
- Explainability is built into the network through secondary outputs for the user

**Abstract**

A common task in forensic biology is to interpret and evaluate short tandem repeat DNA profiles. The first step in these interpretations is to assign a number of contributors to the profiles, a task that is most often performed manually by a scientist using their knowledge of DNA profile behaviour. Studies using constructed DNA profiles have shown that as DNA profiles become more complex, and the number of DNA-donating individuals increases, the ability for scientists to assign the target number. There have been a number of machine learning algorithms developed that seek to assign the number of contributors to a DNA profile, however due to practical limitations in being able to generate DNA profiles in a laboratory, the algorithms have been based on summaries of the available information. In this work we develop an analysis pipeline that simulates the electrophoretic signal of an STR profile, allowing virtually unlimited, pre-labelled training material to be generated. We show that by simulating 100 000 profiles and training a number of contributors estimation tool using a deep neural network architecture (in an algorithm named deepNoC) that a high level of performance is achieved (89% for 1 to 10 contributors). The trained network can then have fine-tuning training performed with only a few hundred profiles in order to achieve the same accuracy within a specific laboratory. We also build into deepNoC secondary outputs that provide a level of explainability to a user of algorithm, and show how they can be displayed in an intuitive manner.

## 1.0 - Introduction

Short tandem repeat (STR) DNA profiles are the most common form of genetic evidence produced in forensic biology laboratories. A standard workflow would involve samples being taken from exhibits of interest to the case, DNA being extracted from those samples, STR regions of the DNA being targeted and amplified by polymerase chain reaction (PCR) and those fragments separated and visualised using capillary gel electrophoresis. The resulting DNA profile is a series peaks present at a number of regions (loci). The profiles are 'read' by analysts, who classify peaks as being artefactual or signal. The profiles can be interpreted and compared to reference DNA profiles (generated from samples taken from individuals of interest in the case) in order to provide information about who is supported as having donated their DNA to the exhibits [1].

The current standard for analysing STR DNA profile data is the use of 'probabilistic genotyping'. This typically involves trialling all possible combinations of reference DNA profiles that could describe the observed evidence profile and (using models for DNA profile behaviour) ultimately assigning a weight to each combination. The weights reflect how well the observed data can be explained. This process is known as deconvolution [2]. Following deconvolution of the evidence profile, a reference profile (from a person of interest) can be compared by considering two competing propositions (that they are a DNA donor, or that they are not a DNA donor to the sample) and a likelihood ratio assigned. This LR is then the probability of observing the evidence profile if the person being compared is a contributor compared to if they are not a contributor [3].

Common to almost all DNA profile interpretations is the need for the analyst to assign a number of contributors (NoC) that they believe have donated DNA to the sample. While there are methods available in probabilistic genotyping tools that allow a range of contributors to be specified [4, 5], even in laboratories that use these tools the most common analysis specifies a single NoC value (authors' personal experience).

The misassignment of number of contributors to a DNA profile can have consequences for down-stream deconvolution. If the NoC is underestimated, then the consequence can be that the true donors of DNA are excluded as contributors to the sample [6]. This occurs as during the deconvolution peaks are forced to pair to create genotypes that do not represent real pairings [7]. If overestimating the NoC then the trend is that non-donors of DNA will tend to be assigned LRs that mildly favour them as being donors to the sample [6]. Neither of these outcomes is desirable and so NoC assignment is an ongoing, and important aspect of DNA profile interpretation to address.

The assignment of a NoC is still usually a manually undertaken process. By manually, we mean that a NoC assignment relies on an analyst examining a profile and using their experience-based knowledge of the performance of DNA profiling in their laboratory. The most common basis for manually assigning NoC is the use of a method known as maximum allele count (MAC), which we elaborate on in section 1.1.1. While this method has a simple basis, there are many complexities that arise as the DNA becomes complex (with greater NoC) or poor quality (such as degraded, low level, or affected by PCR inhibitors).

In addition to the complexity involved in assigning a NoC to a standard DNA profile, analysis of electropherograms is starting to employ the use of neural networks in order to classify the fluorescence into various categories [8, 9]. This system of DNA profile 'reading' no longer requires that peaks in a DNA profile are assigned as either real or artefactual, but instead assigns probabilities for that peak belonging to any one of a number of categories (some artefactual and some not). These probabilities can then be carried forward and used in the deconvolution [10]. From the point of view of assigning a NoC the task becomes even more difficult in this context as the analyst must take into account multiple peaks that may only have a low probability of being non-artefactual when making the assignment.

The assignment of a NoC is one of the few remaining barriers to automation of DNA profile analysis. In the work by Taylor et al [10] where the authors show the use of artificial neural networks to automatically process electrophoretic DNA profile data, and then pass the resulting output into a probabilistic genotyping deconvolution. They state

> *The current workflow used in this paper still required the assignment of the number of contributors. The automation of this task is the final step that would be required in order for a workflow, that was able to be completely automated from laboratory to deconvolution.*

This all points to the need for a tool that can use the information within an EPG and assign a NoC (or provide a probability distribution across a range of plausible NoC). To assist in assignment of NoC there have been a number of machine learning tools developed, which we summarise in section 1.1. The most advanced of these are those that rely on some type of machine learning, either decision tree [11], a random forest algorithm [12], or a multi-layer perceptron deep learning model [13]. However, there are limitations with these currently available systems. Each system is trained on DNA profiles constructed from known donors in a laboratory. This limits (due to cost and practicality) the number of training profiles that can be accessed. There are large online datasets of profiles available [14], but even this is limited to a few thousands of profiles per profiling system (most of which are from a single DNA contributor), and limited to a complexity of five contributors (well below what is encountered in forensic casework, for example see experiment 2 in [15]). Due to these dataset limitations the machine learning algorithms are trained on data that has been simplified by:

- filtering to remove artefacts,
- removing any potential peaks that occur below an analytical threshold,
- often only using high level summaries of features of the DNA profiles,
- Limits the complexity of the classifications to 3, 4 or 5 contributors (which is acceptable for comparison to the constructed mixtures that are similarly limited in this way, but would cause issues when provided casework samples that exceed this complexity).

Each of these points reduce the information content within the profile and thereby reducing the ability for a machine learning system to learn to assign a NoC.

Our work seeks to develop a deep learning NoC classifier that differs from previous work in several ways. We don't:

- filter artefacts or stutters from the profiles,
- apply an analytical threshold to screen out low intensity peaks, or
- provide only summaries of the profile data.

Instead, we start with the raw electrophoretic signal and use it to train a deep learning neural network, deepNoC. Given the complexity of this data it is likely to require a large training dataset to achieve high accuracy, well beyond the practical limits of a laboratory or online repository. To overcome this issue we utilise a recently published generative adversarial network (GAN) [16] to simulate the electrophoretic signal of 100 000 GlobalFiler™ PCR Amplification Kit DNA profiles (kit designed by Thermo Fisher Scientific) that range from 1 to 10 contributors. We develop a pipeline of analysis from simulating profiles, detecting peak information, assigning peak classification probabilities, preparing deepNoC input data and automatically labelled output data (possible as the profiles were simulated).

In addition to training deepNoC to a high level of performance, we concentrate on ensuring the assignment made by deepNoC are explainable. In artificial intelligence (AI) development generally there is an increased focus on developing explainable AI (XAI) tools. Governments worldwide are developing regulations on the application of AI in high-risk settings (such as forensic science) and white papers on XAI (for example [17]). There have also been recent works in forensic biology showing the application of XAI to machine learning algorithms sed to assign a NoC [18], and when classifying fluorescence in an EPG [19]. In deepNoC we build explainability into the structure of the neural network (NN) by including outputs that give insight into how each peak, and each locus is being considered by the algorithm. This has an advantage over other XAI methods as it does not need to be a post-hoc applied calculation, the explainability is given to the user at the same time as the NoC estimation.

## 1.1 - Current methods for assigning NoC

### 1.1.1 – Maximum Allele Count (MAC)

The most basic form of assigning a NoC is the maximum allele count (MAC), which relies on counting the number of alleles at each locus, finding the maximum and using the knowledge that one person can only donate up to two alleles per locus, assigning a NoC (where 1 or 2 peaks could come from one person, 3 or 4 peaks could come from 2 people, etc). Although not specifically studied, it is widely accepted than manual assignment of NoC fundamentally carries out a MAC process, but then also takes into account the peak heights to further refine the assignment. As DNA profiles become more complex (i.e. originating from DNA from large numbers of contributor) then there is a greater level of allelic overlap (i.e. the same alleles donated by more than one donor) [20] and human performance in assigning the NoC decreases [21].

### 1.1.2 - Probabilistic Assessment for Contributor Estimation (PACE)

Marciano et al [22] initially trialled k-nearest neighbours, classification and regression trees, multinomial logistic regression, multiplayer perceptron and support vector machine algorithms. They later used an ensemble method utilising a random forest algorithm [12]. The features they used were a series of summaries of the profile (MAC, total peak count, peak heights, peak height rations etc) based on the profile that had been pre-processed to remove artefacts. On the GlobalFiler™ DNA profiling system, for profiles that ranged from one to four people they achieved approximately 90% accuracy.

1.1.3 - TAWSEEM

Alotaibi et al [13] developed TAWSEEM (from the Arabic word meaning 'labelling') which was a deep learning multi-layer perceptron with 15 hidden layers (each with 64 neurons). They applied their model to the ProvedIt dataset profiles ranging from 1 to 5 contributors in several ways, either as data from single multiplexes or combining data from up to four multiplexes. In the largest of the training sets (utilising data from four multiplexes) they trained their model on 6000 profiles. They ultimately achieved an accuracy of 89% on a single-multiplex model and an accuracy of 97% on their dataset, although the classifications appear to be based on those of individual loci rather than whole profiles and it is not clear how these relate to whole profile results.

1.1.4 – Decision Tree

Kruijver et al [11] developed a decision tree which used profiles from the ProvedIt dataset, with stutter peaks filtered and an analytical threshold of 10rfu. The trees they trained used covariates relating to the number of alleles in the profile, and loci, some peak height related summaries and an indication of profile rarity. They trained the model on 300 profiles ranging from 1 to 5 contributors and tested the performance on 466 profiles. They achieved an overall accuracy of 77% across the test dataset.

1.1.5 – Total Allele Count

Noel et al [23] modelled the total number of alleles per profile for 2 to 7 person mixtures and created total allele count (TAC) distributions for each NoC. The model was trained on simulated data, which was complete (i.e. not weak, or with any missing information) and without artefacts or stutters. They tested their system on a set of 21 IdentiFiler profiles ranging from 3 to 5 contributors from the ProvedIt dataset, but do not provide performance statistics. For 81 simulated IdentiFiler profiles they obtained between 100% (for 3-person mixtures) to 65% (for 6-person mixtures) correct assignments.

1.1.6 – Maximum likelihood estimate (MLE)

Haned et al [24] investigated the ability to use the maximum likelihood estimation, based on the frequency of the alleles present in a profile. They tested the model on 1000 simulated mixtures ranging from 2 to 5 contributors and achieved around 65% accuracy in assignment.

1.1.7 – Trans dimensional Markov Chain Monte Carlo

Taylor et al [4] adapted a method of Weinberg et al [25, 26] to calculate a Bayes Factor from a Markov Chain Monte Carlo system and apply it to a probabilistic genotyping DNA profile deconvolution tool. In doing so they were able to analyse a profile under different NoC and use the Bayes Factor to weight the support of the model for each NoC in the range tested. McGovern et al later testing this on a range of profiles [5]. While this could be used as a NoC assignment tool, the fact that the profile has to be deconvoluted under each NoC means that it can take considerable time for a complex analysis to run. The authors designed the tool as a means of combining the individual deconvolutions (with different NoC) into a single evaluation, thereby treating the NoC as a nuisance variable, and so there was no explicit comparison of true NoC to favoured NoC using the tool.

1.1.8 – Random forest

Benschop et al [27] trialled a series of 10 machine learning algorithms on a set of 590 PowerPlex® Fusion 6C profiles ranging from 1 to 5 contributors. They found the best performing algorithm was a random forest classifier that utilised 19 features from the DNA profile. Using this model they achieved 83% accuracy across the test set. The same group later applied explainability tools, Shapley values and counterfactuals to the random forest model to demonstrate which features were important in decision making [18].

1.1.9 – Posterior probability

Swaminathan et al [28] developed a method called NOC*It* that calculated the posterior probability of the number of contributors based on the peaks observed in the DNA profile, their height and their population frequency. Their model does not screen out stutter peaks and can deal with missing peaks that have 'dropped out'. They tested their system on 278 IdentiFiler® Plus laboratory generated DNA profiles ranging from 1 to 5 contributors and achieved 83% of samples.

**2.0 - Method**

2.1 – Pipeline of data simulation

DNA profile peak information (i.e. information about peak locus, allelic designation, height, and size) were simulated using R package simDNAmixtures [29]. These profiles were simulated to be from 1 to 10 contributors and calibrated to the DNA profile behaviour of GlobalFiler as produced under laboratory conditions at Forensic Science SA (based on the calibration information used to set up probabilistic genotyping software STRmix [30] in their laboratory). Each contributor could contribute an amount of DNA that would render their peaks virtually indistinguishable from baseline noise, up to a level close to saturation. This profile peak information was used to generate simulated EPG signal using the generator neural network from the GAN outlined by Taylor et al in [16].

Next peak centres were detected in the simulated EPG using the method of Woldegebriel [31] and labelled with their locus, allelic designation, size, and height. The majority of the simulated peaks from simDNAmixtures were relabelled in this process, however there were low level peaks that were not relabelled, due to being obscured once by the 'instrument noise' introduced by the conversion to an EPG. There were also many instances of additional peaks being created when the EPG was simulated from peak information, as the creation of the EPG introduced peak artefacts such as pull-up.

The EPG data from around the peak centres was classified by the multi-head convolutional neural network (MHCNN) described in Taylor et al [9] in order to also assign a 'peak label probability (plp)' to each peak i.e. a probability that the peak was non-artefactual (defined as being allelic or stutter).

This completed the simulation of the DNA profiles to be used as input to deepNoC, but there were three pairs of outputs also generated during the simulation process. The first pair of outputs related to the peak level and provided the number of alleles that were in that allelic position (i.e. anywhere from 0 to 2×NoC) and the proportion of the peak that was allelic. The second pair of labelled outputs was at the locus level and provided the number of alleles at the locus (always 2×NoC) and the proportions that the contributors were present at the locus. The third labelled output was at the profile level and provided the proportions that the contributors were present in the whole profile, and the NoC.

2.3 – deepNoC architecture

Figure 1 shows the basic architecture of deepNoC. A structure with more detailed information is provided in the supplementary material.

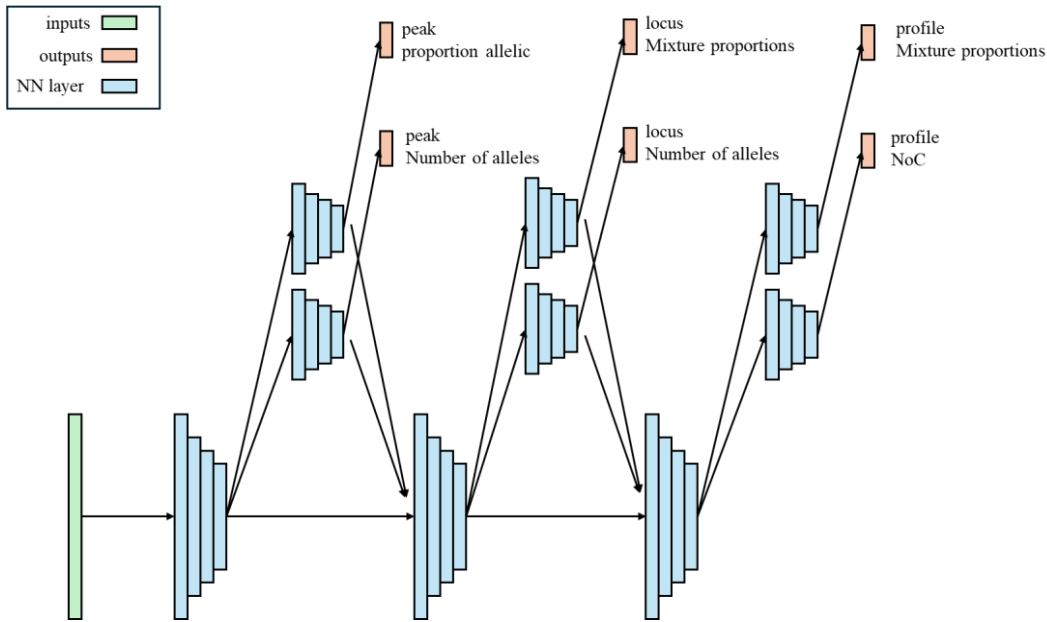

*Figure 1: deepNoC architecture.*

From the input to the profile NoC output there are 16 layers. The input for each profile is [24 × 50 × 89] which we describe in more detail in section 2.4. In addition to the 16 layers between input and profile NoC there are additional layers that produce secondary outputs (at the peak, locus, or profile level). These secondary outputs are designed to provide explainability to the NoC assignment of the model. The values of the secondary outputs are also fed back into the main branch of the NN to assist in further learning. A variant of the architecture seen in Figure 1 was trialled that had no secondary structures (i.e. the input simply passed through 16 layers of the main branch of the NN to the profile NoC, with no other outputs) to ensure that the performance of the model did not suffer from the additional outputs. Trials showed no noticeable difference in overall performance (data not shown) and so the architecture seen in Figure 1 was used, as it had the advantage of explainable outputs.

2.4 – Input data structure

Each profile input was [24 × 50 × 89] where the first dimension represented the 24 loci in the GlobalFiler kits, the second dimension represented the ability to capture up to 50 peaks at each locus and the third dimension represented the 89 pieces of information collected about each peak. If there were any locus with no peak information, then the array was filled with zeros. Similarly, if there were less than 50 peaks the additional peak positions were filled with zeros. If there were more than 50 peaks at a locus then only data for the first 50 peaks was captured.

To reduce the number of peaks at each locus to only those that had the most impact on the assignment all peaks with a probability greater than 0.97 of being artefactual were removed from the inputs (as was done in [10]).

The 89 inputs for each peak were:

- 1 to 24: Locus – one-hot encoded locus (i.e. a 1 and 23 zeros)
- 25: the allele designation ( / 100)
- 26: The size in base pairs ( / 100)
- 27: The height in rfu ( / 33 000)
- 28: The allele frequency in the Australian Causcasian population [32]
- 29: The peak label probability

If the peak is stutter peak of type $i$ then:

- 30: the parent allele designation ( / 100)
- 31: the parent peak height rfu ( / 33 000)
- 32: the ratio of peak height to parent height (capped at 1)
- 33: the expected stutter ratio of a stutter in this peak position
- 34: the frequency of the parent allele
- 35: the peak label probability of the parent allele

Type $i$ was varied as: a back stutter (data point 30 to 35), a double back stutter (points 36 to 41), a forward stutter (42 to 47), a point 2 repeat (48 to 53).

If the peak is a parent peak of stutter type $j$ then:

- 54: the stutter peak allelic designation ( / 100)
- 55: the stutter peak height rfu ( / 33 000)
- 56: the ratio of stutter peak height to peak height (capped at 1)
- 57: the expected ratio that a parent allele in this position is expected to stutter
- 58: the frequency of the stutter allele
- 59: the peak label probability of the stutter peak

Type $j$ was varied as: a back stutter (54 to 59), a double back stutter (60 to 65), a forward stutter (66 to 71), a point 2 repeat (72 to 77).

- 78: The total peaks at the locus ( / 100)
- 79: The total peaks in the profile ( / 1000)
- 80 - 89: The expected mixture proportion of the 1$^{st}$ to 10$^{th}$ largest contributing individual if there were 10 donors

The mixture proportions making up data points 80 to 89 were obtained by passing the input profile through a previously published algorithm (see Appendix 1 of [15]) called the 'smart start' algorithm, used for assigning DNA template starting amounts in probabilistic genotyping software STRmix. If the algorithm is asked to assign mixture proportions to more contributions than what is needed to explain the DNA profile, then the superfluous contributors are assigned a small default non-zero DNA amount.

Point 78 is going to be the same for all peaks within a locus, and points 79 to 89 are going to be the same for all peaks in the DNA profile, meaning that there is some repetition in the input.

The labelled outputs for each profile were in the format:
- 1. Peak proportion allelic: $[24 \times 50 \times 1]$ with the values lying between 0 and 1
- 2. Peak number of alleles: $[24 \times 50 \times 21]$ with the output for each peak being a one-hot encoded array representing the number of alleles in contributors with that designation, thereby ranging from 0 to 20
- 3. Locus mixture proportions: $[24 \times 10]$ with the final dimension representing the proportion of each donor to the locus. Each array of 10 values sums to 1.
- 4. Locus number of alleles: $[24 \times 20]$ with the output for each locus being a one-hot encoded array representing the number of alleles in contributors with that designation, thereby ranging from 1 to 20
- 5. Profile mixture proportions: $[10]$ representing the proportion of each donor to the sample. Each array of 10 values sums to 1.
- 6. Profile NoC: $[10]$ one-hot encoded array representing the number of contributors.

Each of these outputs could be constructed from the simDNAmixture object created when a profile is simulated.

2.5 – Training on simulated data

Training was conducted using a batch size of 100 and carried out for 200 epochs. The 100000 simulated profile dataset was split into 90% training and 10% test datasets. Training was carried out using python V3.10 and TensorFlow V2.10 (GPU) with CUDA V11.8.89. An Adam optimiser was used with learning rate 0.00001 and a beta parameter of 0.5. The loss functions for the 6 outputs listed in section 2.4 were mean squared error (outputs 1, 3 and 5) and categorical cross-entropy (outputs 2, 4 and 6).

2.6 – Fine-tuning training

To test the performance of the model on laboratory-created profiles, fine-tune training was carried out on GlobalFiler profiles from the ProvedIt archive that had been generated on an ABI 3500 electrophoresis instrument, using a 25 second injection time. A decision was made not to use GlobalFiler profiles from multiple methods due to the fact that this would not represent the way in which deepNoC could be used and trained in a forensic laboratory (which will only be using a single laboratory method and whose capacity to generate DNA profiles for training and validation will be limited). This led to 743 profiles that ranged from 1 to 5 contributors. The number of single source profiles available vastly outweigh the number of mixtures on the ProvedIt archives and so not all single source profiles were included in the fine tuning. The final result was 68 (1-person), 175 (2-person), 158 (3-person), 186 (4-person), 156 (5-person) profiles. For fine tune training every second profile was used to train (371 profiles) and the remaining (372 profiles) were used to test. This number of training profiles (i.e. 370) is likely to be at the limit of what a typical forensic laboratory can reasonably produce for validation (author personal experience).

All outputs could be created as the profiles were constructed and the references of the donors were known. The only output that was not able to be easily created was the proportion of each peak that was allelic due to the fact that random stochastic effects within laboratory-generated profiles can never be known. For these values the profiles were passed through the deepNoC model once prior to fine-tuning and the predicted peak output for proportion allelic was taken and then used as the labelled output in the fine-tune training.

2.7 – Explainability output

A method was written to display the results of all outputs graphically in one image and overlay the DNA profile. This was done so that a scientist using deepNoC could assess the performance of the model by the additional information from the XAI components built into the structure of the model.

2.9 – Computational tools

The analysis was performed on an Intel® Core™ i9-14900HX running at 2.20 GHz with 128GB RAM and a 16GB Nvidia RTX4090 GPU. The PC was running Windows 11 Professional V32H2.

**3.0 - Results**

3.1 – description of simulated data

The simulated dataset consisted of a wide range of template DNA amounts and total number of peaks for all NoC categories (Figure 2). These profiles are likely to cover the range of complexity and makeup encountered in casework and exceed the level which would be carried through to analysis (which for many laboratories is a limit of four contributors). Figure 2 shows that as the NoC exceeds 7 there is minimal increase in the number of peaks in the profile. This is expected to make classification of NoC between these higher categories particularly challenging for any NoC assignment tool. The expected total template increases linearly with each additional NoC (which is expected given the simulation setup), however across all categories there is a wide overlap in the distribution of total DNA.

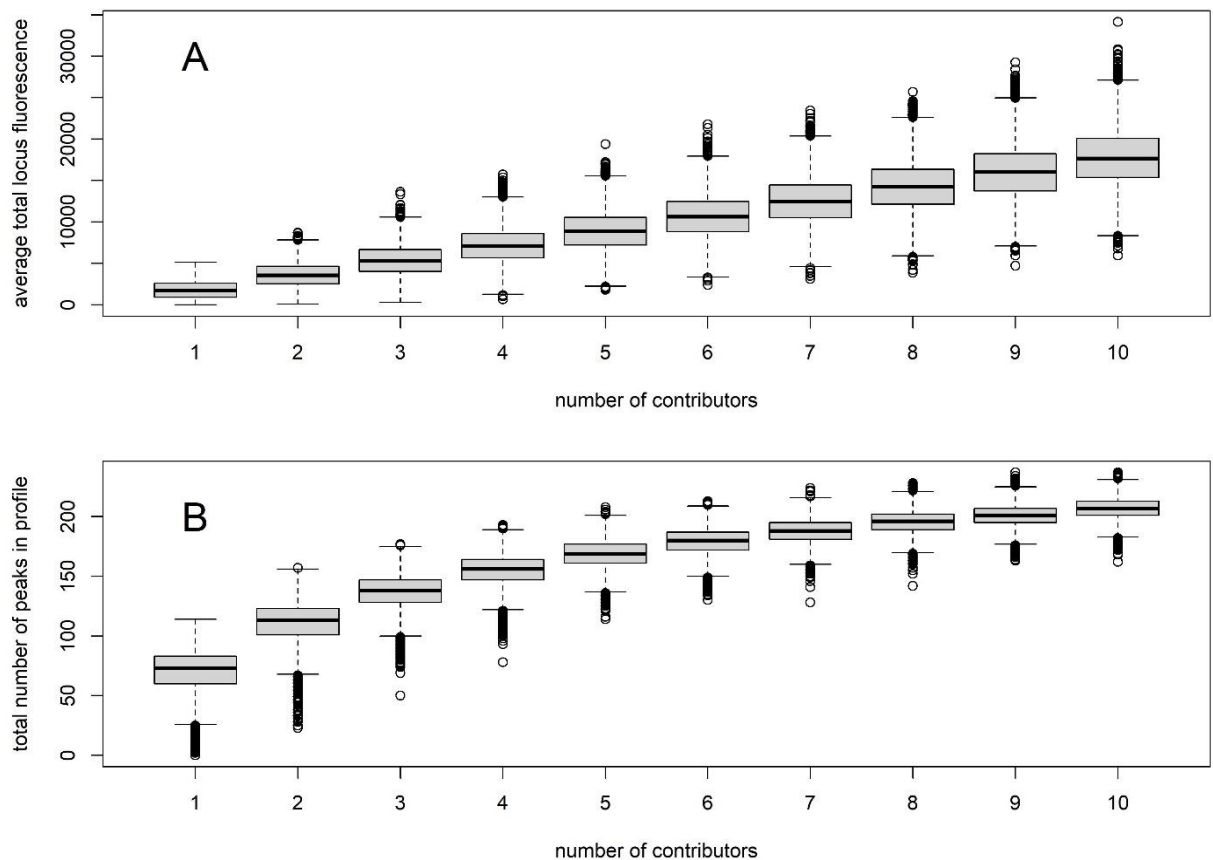

*Figure 2: distribution of total DNA (A) and number of peaks (B) for each NoC within the simulated profile set*

3.2 – Performance on simulated data

When considering the full 100 000 dataset (10 000 test and 90 000 training) close to the maximum accuracy of the model on the test set was reached by approximately 50 epochs, with only gradual improvement after that (Figure 2). Additional training provided no further improvement in accuracy for the test set (data not shown).

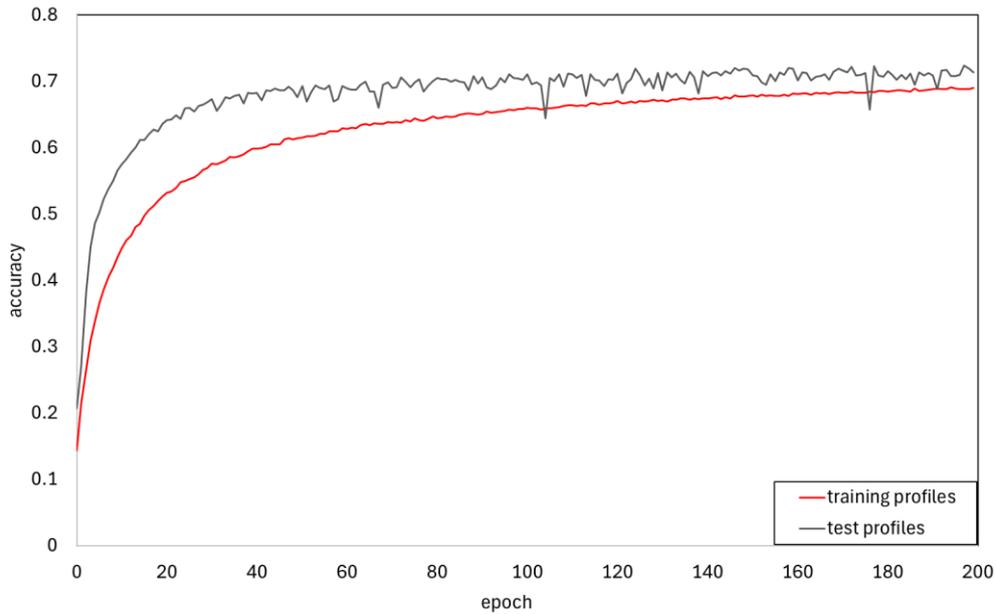

*Figure 3: Traceplot showing accuracy in NoC assignment for training and test sets over 200 epochs of training*

Table 1 shows the performance of deepNoC in assigning NoC for the 10000 profiles in the test set. As expected, the performance of the classifier is best for simpler profiles (i.e. those that have originated from a single, or two contributors) and as the NoC increases, the classifications become more spread over a number of possibilities (but always with the highest number of classifications being made into the ground truth category). There is a steady decline in accuracy as NoC increases, except for the final category, a NoC of 10, which sees a jump in accuracy again compared to a NoC of 9. This jump is due to the fact that when dealing with profiles coming from the final category the assignments become easier. Any profile of a certain complexity will be assigned to this last category and is likely to be correct (knowing that in the dataset there can be no higher number than this). This 'final category skew' of accuracy is seen in various other published model performance (which we discuss later).

|   |   | \multicolumn{10}{c}{**known NoC**} |   |
|---|---|---|---|---|---|---|---|---|---|---|---|
|   |   | 1 | 2 | 3 | 4 | 5 | 6 | 7 | 8 | 9 | 10 | Total |
|   | 1 | 983 | 0 | 0 | 0 | 0 | 0 | 0 | 0 | 0 | 0 | 983 |
|   | 2 | 2 | 1020 | 1 | 0 | 0 | 0 | 0 | 0 | 0 | 0 | 1023 |
|   | 3 | 0 | 41 | 948 | 13 | 0 | 0 | 0 | 0 | 0 | 0 | 1002 |
| Predicted NoC | 4 | 0 | 0 | 149 | 811 | 41 | 0 | 0 | 0 | 0 | 0 | 1001 |
|   | 5 | 0 | 0 | 3 | 198 | 704 | 94 | 1 | 0 | 0 | 0 | 1000 |
|   | 6 | 0 | 0 | 0 | 10 | 219 | 622 | 113 | 2 | 0 | 0 | 966 |
|   | 7 | 0 | 0 | 0 | 0 | 21 | 272 | 489 | 206 | 8 | 0 | 996 |
|   | 8 | 0 | 0 | 0 | 0 | 1 | 34 | 243 | 545 | 184 | 38 | 1045 |
|   | 9 | 0 | 0 | 0 | 0 | 0 | 2 | 40 | 298 | 367 | 285 | 992 |
|   | 10 | 0 | 0 | 0 | 0 | 0 | 0 | 4 | 91 | 249 | 648 | 992 |
|   | Total | 985 | 1061 | 1101 | 1032 | 986 | 1024 | 890 | 1142 | 808 | 971 | 10000 |

*Table 1: Confusion matrix showing performance of model on 10,000 simulated profiles in test set*

The size of the training dataset had a clear relationship with the performance of the classifier (Figure 4). Even training the deepNoC model with only a few hundred profiles (the number of profiles that a typical forensic laboratory might produce for carrying out a validation) achieved around 40% accuracy, showing some power in even limited data for training. As the amount of training data increased to 90000 the accuracy increased to the 72% obtained in the final model (with results seen in Table 1). At this point the accuracy was still clearly rising with increased number of training profiles, and it may have been possible to increase the accuracy to closer to 80% (if the trend continued) with the use of around 1 million simulated profiles. Simulation was halted at 100 000 profiles for two reasons; a) the availability of computing power and b) the limited effect of additional accuracy on simulated profile data this would have when it came to applying the trained model to laboratory-generated profiles.

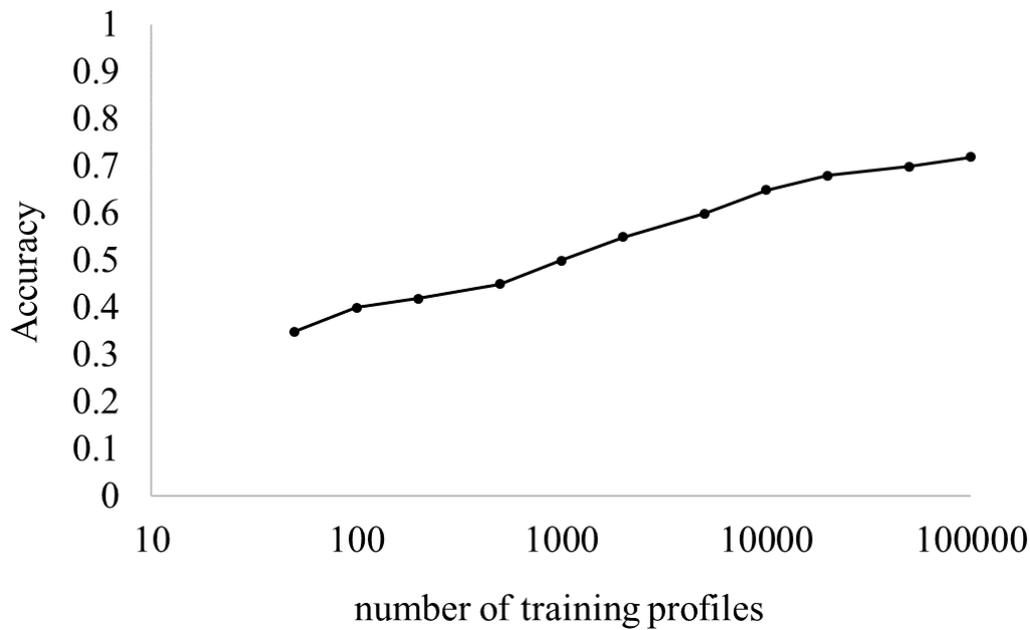

*Figure 4: Learning curve showing accuracy of deepNoC classification on 10 000 simulated profiles in test set when trained on 50 to 90 000 simulated profiles*

3.3 – Performance after fine tuning training

Figure 5 shows the traceplot for the fine-tune training performed on the ProvedIt laboratory-generated GlobaFiler profiles. It can be seen in Figure 5 that the performance on laboratory-generated data initially was around 40% to 60%, which is an expected drop from the 72% accuracy obtained on simulated test data. However, the performance quickly rose to around 90% on the test set after 200 epoch and remained at this level. The improvement in accuracy on the laboratory-generated data compared to simulated data is likely to result from two effects. First, the simulated dataset possessed profiles with a larger range of NoC (and specifically higher numbers, which is where the accuracy is lowest). Second, the profiles produced by the simulation can produce profiles that are more difficult than would typically be produced in a laboratory for a validation (as validation profiles tend not to concentrate on producing many profiles that are expected to be beyond the limits of interpretation).

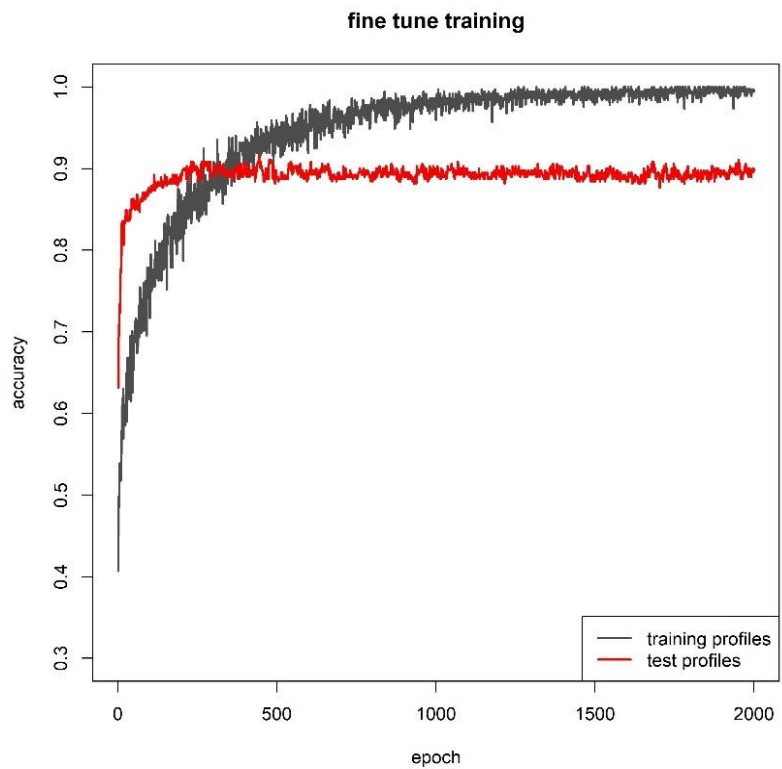

*Figure 5: Traceplot showing accuracy of deepNoC on ProvedIt dataset*

Table 2 shows the outcome of 2000 epochs of fine-tune training on the test set of ProvedIt profiles. As with the simulated profiles, the performance of deepNoC is highest when profiles originate from fewer individuals. Table 2 shows high metric values for accuracy, precision, recall and F1 score for all NoC. We note that there were no classifications of any profiles beyond 5. This may have been a feature learned by deepNoC during fine-tune training i.e. that profiles will not appear that come from more than 5 contributors, and therefore an earlier stopping on the stopping may be preferred to prevent this (if it is indeed occurring).

|  | | known NoC | | | | | Accuracy | Precision | Recall | F1 |
|---|---|---|---|---|---|---|---|---|---|---|
|  | | 1 | 2 | 3 | 4 | 5 |  |  |  |  |
| Predicted NoC | 1 | 34 | 0 | 0 | 0 | 0 | 1.000 | 1.000 | 1.000 | 1.000 |
|  | 2 | 0 | 88 | 2 | 1 | 0 | 0.992 | 1.000 | 0.967 | 0.983 |
|  | 3 | 0 | 0 | 69 | 2 | 1 | 0.965 | 0.873 | 0.958 | 0.914 |
|  | 4 | 0 | 0 | 6 | 82 | 16 | 0.911 | 0.882 | 0.788 | 0.832 |
|  | 5 | 0 | 0 | 2 | 8 | 61 | 0.927 | 0.782 | 0.859 | 0.819 |

*Table 2: confusion matrix for deepNoC assignments on ProvedIt test dataset and associated performance statistics per NoC category*

As well as considering the category that had the maximum probability (and to which the input was assigned) the distribution of probabilities across all possible categories can be viewed. Figure 6 shows the distribution of probabilities, and for the majority of profiles the highest probability is close to 1. The exception is for misassigned profiles, where there tends to be a gradation of probability assigned to the correct category to the incorrect category.

In forensic science, it is common that analyses are performed only once a level of confidence has been attained e.g. that a number of contributors can be assigned with enough confidence. A level of information wastage (i.e. profiles that are not interpreted) is accepted to maintain this confidence threshold. The same thinking can be applied to the NoC classification output of deepNoC. Rather than every input being classified, a probability threshold could be set so that a classification only occurs when the maximum probability meets that threshold. This results in each profile either being classified to a NoC category or being unclassified. Figure 7 shows the trade-off between the classification accuracy and the proportion of profiles that would be classified over a range of threshold values (based on the ProvedIt dataset). An accuracy of close to 100% could be achieved if a laboratory was willing to have approximately half of their profiles unclassified. However, the most rapid drop in proportion of classified profiles does not occur until the highest thresholds. An accuracy of 95% could be achieved whilst still classifying approximately 90% of profiles using a classification threshold of approximately 0.95.

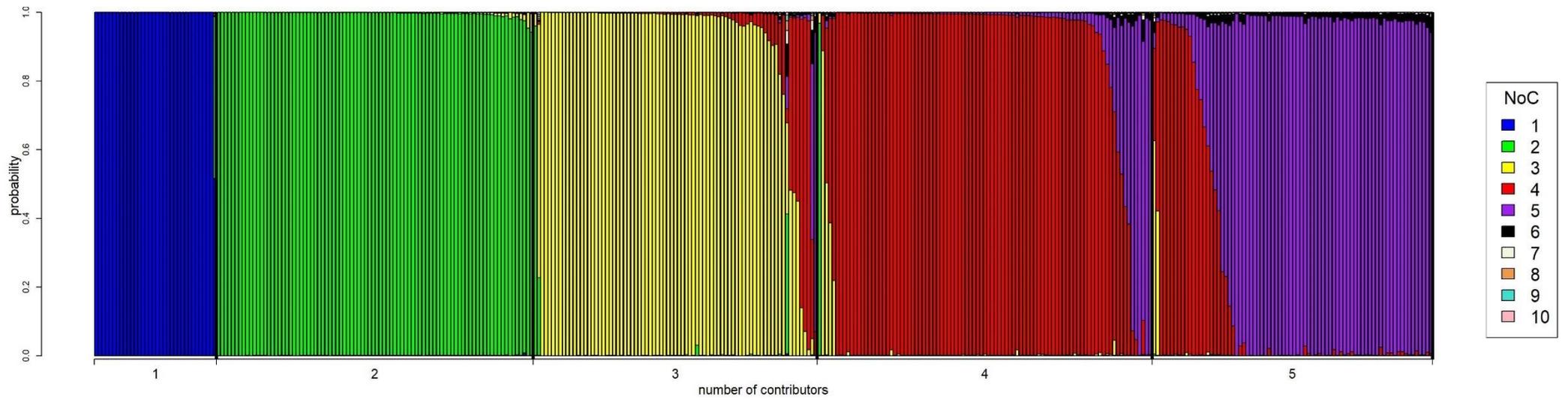

*Figure 6: Assignment probabilities for the 370 test ProvedIt profiles into available NoC categories in deepNoC model*

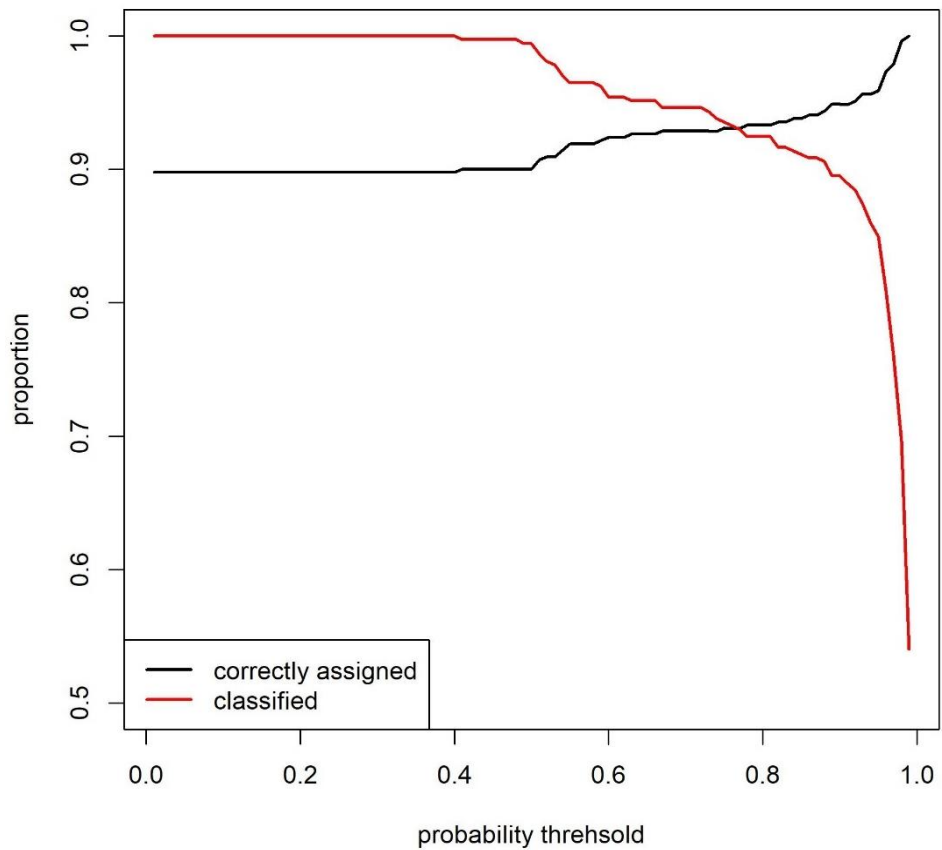

*Figure 7: Assignment accuracy of 370 ProvedIt test profiles (black line) and proportion of profiles assigned (red line) when using minimum probability assignment threshold*

3.4 – Explainability output

The multiple outputs of the ANN shown in Figure 1 can be used as points of explainability. As well as providing the deepNoC classification result, an R script was written to take the input file and the outputs of deepNoC to produce a summary image that can be used by analysts to assist with explainability. An example of the explainability output is shown in Figure 8. The input file is used to draw an idealised version of the peaks (in dye colours of the GlobalFiler profiling kit), signifying the bounds of each locus. A grey barplot sitting behind each peak shows the allele probabilities. All these mentioned components come from the profile input. Each peak has outputs that relate to the number of alleles that are expected to be present in that peak position, and the proportion of that peak which is allelic. On the output shown in Figure 8 the proportion of the peak that deepNoC expects as being allelic is shown by a red bar intersecting the peak at the estimated proportion. The distribution of probability across the number of alleles for each peak was not displayed on the output for two reasons; a) it created a visually clustered image and b) the locus level number of alleles provided more meaningful information for interpretation.

The number of alleles expected on a locus-by-locus basis is show within the bounds of each locus as is the expected mixture proportions. Finally, at the profile level (on the right of the image) the overall mixture proportion and distribution of probability across the NoC are given, with the classified NoC being highlighted in red.

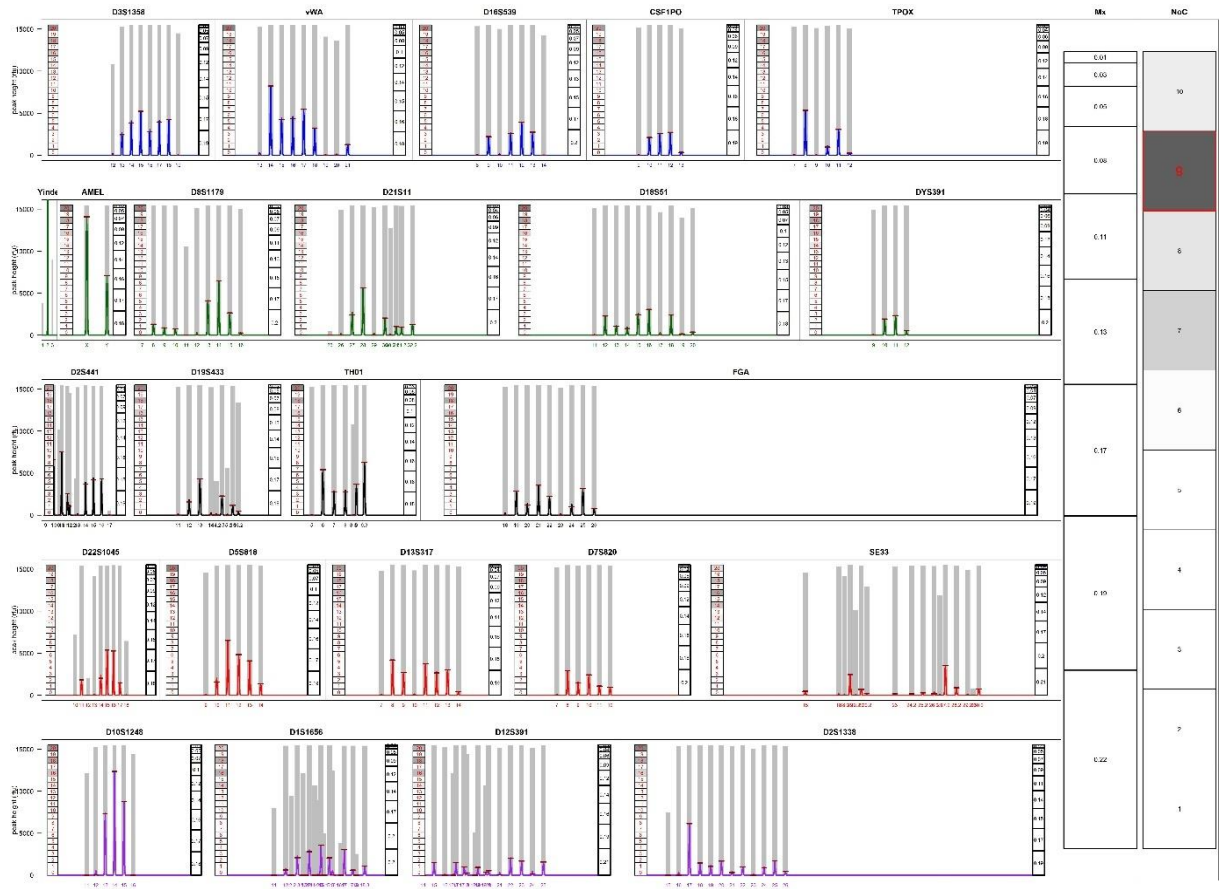

*Figure 8: Example of explainability output produced from deepNoC output and input profile*

**4.0 - Discussion**

4.1 – comparison to other works

The ProvedIt dataset [14] is a common set of publicly available profiles that have been used in a number of NoC classification trials. This allows a direct comparison of the performance of deepNoC against other algorithms. In Table 3 the accuracy of NoC classification algortihms per NoC is shown. The top row shows the performance of deepNoC on the simulated profile dataset up to 10 contributors. The middle row of the table shows the performance of deepNoC on the ProvedIt dataset, alongside the performance of several other existing methods. The MAC, Decision Tree, NoCIt, and LDA-26 accuracy values were all obtained from Kruijver et al [11]. The results for PACE-GF were taken from Marciano et al [12]. The results for TAWSEEM were taken from the single multi-plex results of Alotaibi et al [13]. It appears that the Alotaibi et al results capture the classifications on a per-locus basis rather than a per-profile basis. It is not clear how this is done i.e. classifications are provided for each locus individually, or whatever NoC the profile is classified as coming from, then this is multiplied by the number of loci in that profile with data and tallied in their confusion matrix. We suspect the latter given that there is no discussion in their paper on how conflicting classifications at different loci are handled. If the latter method is used, then it might skew results to appear to have a higher accuracy than other algorithms that capture results on a per-profile basis (as all others do). For example, imagine two profiles, one strong with data at 24 loci and one weak with data at only two loci. The strong profile is likely to be correctly classified and the weak profile is more likely to be misclassified (due to little information). TAWSEEM would count this as 24 correct and 2 incorrect classifications whereas other methods in Table 3 would count this as 1 correct and 1 incorrect. Regardless of this point we calculate the accuracies based on the locus tallies provided in their publication. We also note that the TAWSEEM method showed higher accuracy on their multiple-multiplex model, however the results for this analysis train on different systems to the GlobalFiler kit and so are no longer directly comparable. It is also not clear how a laboratory would train their own data in a multi-multiplex framework. Hence we have used the single-multiplex model results.

As mentioned earlier in the paper, when a system is trained to classify NoC then there is a tendency for the highest category in the range to have an inflated accuracy. This can be seen clearly when there is a consistent decrease in accuracy as NoC increases, except for the last category which will exhibit an accuracy increase again. In Table 3 categories that have potentially skewed accuracies have been highlighted in red and are not considered in the determination of the highest result per NoC. The highest performing method on per NoC basis is highlighted in green, which for the majority of categories is deepNoC. While the three deep-learning tools (deepNoC, PACE-GF, and TAWSEEM) are all close in performance, deepNoC has the advantage that it is not limited to laboratory-generated profiles for training. Hence it is likely to be useful in practice where there will be many profiles provided to it from casework that have greater than 5 contributors.

The final line of results in Table 3 shows the performance of human assignments. These results are obtained from Bright et al [21] and relate to a large inter-laboratory study that was conducted using multiple different profiling systems. In the paper the authors requested profiles that appeared as originating from three to five people. Therefore, six-person mixtures would only be provided if they looked like they originated five, four or three people. However, this requirement is not likely to have screened out a large portion of six-person profiles. Coble et al [33] showed that for the GlobalFiler system between 86% and 93% of complete six-person mixtures would have a number of peaks (not taking into account their balance) which could be described by five people or less. At the lower levels of three-person mixtures there is a chance that the accuracy is skewed as any three-person mixtures that appeared as two-person mixtures would also have been excluded from the data capture.

|  |  | accuracy | | | | | | | | | |
|---|---|---|---|---|---|---|---|---|---|---|---|
|  |  | 1p | 2p | 3p | 4p | 5p | 6p | 7p | 8p | 9p | 10p |
| **Simulated data** | deepNoC | 0.9998 | 0.996 | 0.979 | 0.959 | 0.942 | 0.925 | 0.909 | 0.890 | 0.893 | 0.933 |
| **ProvedIt GlobalFiler data [14]** | deepNoC | 1.000 | 0.992 | 0.965 | 0.911 | 0.927 | | | | | |
|  | PACE-GF | 0.986 | 0.963 | 0.944 | 0.959 | | | | | | |
|  | TAWSEEM | 0.979 | 0.951 | 0.941 | 0.937 | 0.969 | | | | | |
|  | MAC | 0.627 | 0.806 | 0.84 | 0.583 | 0.087 | | | | | |
|  | Decision Tree | 0.910 | 0.960 | 0.819 | 0.602 | 0.660 | | | | | |
|  | NoCIt | 0.806 | 0.939 | 0.84 | 0.825 | 0.573 | | | | | |
|  | LDA-26 | 0.925 | 0.970 | 0.904 | 0.825 | 0.806 | | | | | |
| **Varied real profile data** | human | | | 0.915 | 0.618 | 0.768 | 0.750 | | | | |

*Table 3: Accuracy of deepNoC model on simulated test set (top row), accuracy of various available machine learning tools for assigning NoC when applied to ProvedIt dataset (middle row) and accuracy of human assignment based on various profiles from [21] (bottom row). Cells highlighted green are the highest accuracy for that NoC within the methods trialled on the ProvedIt dataset. Cells highlighted red = may be skewed by experimental design (and not counted in assignment of the highest accuracy model)*

4.2 – Future directions

4.2.1 – extending to accommodate multiple PCRs

A common practise in forensic biology is for the same sample to be used multiple times to generate replicate DNA profiles. The combined information from these replicates provides greater ability to interpret the DNA profile, assign a NoC and distinguish contributors from non-contributors [7]. The deepNoC structure developed in this paper works on only a single PCR replicate. However, there is an opportunity for the multiple replicates to be taken into account during a machine-learning-algorithm-based assignment of NoC. This could be done in several ways:

- A structure that works on a single replicate could be maintained and when multiple replicates are available the tool is used on each individually and the results combined afterwards in some way to assign an overall NoC. For example, a single method could be to simply take the largest NoC assigned across the multiple replicates. Another option could be to combine the probabilities for each NoC across replicates and normalise to obtain an overall value for assignment (this could work particularly well in conjunction with a posterior probability threshold to screen out profiles where the two replicates strongly disagreed in NoC).
- Additional datapoints could be added to the final dimension of the input i.e. instead of [24 x 50 x 89] for a profile it could be [24 x 50 x 178] (where 178 is twice 89 but noting that not all datapoints would need to be duplicated as some are not peak specific). Then for problems that have a single PCR the latter half of datapoints are assigned a value of 0. For problems that have two PCRs then the full complement of information is used.
- A separate neural network could be trained for multi-PCR problems. This would allow the input for a profile problem to have an additional dimension to reflect the peak properties in each replicate i.e. an input would be [24 x 50 x 89 x 2] for a system that was trained on two-PCR problems and the initial layers would be 3D CNN.

4.2.2 – Extending the explainability

Part of the structure when creating deepNoc was to incorporate explainability into what the system learned. This was facilitated by the use of multiple outputs at the peak, locus or profile level. While useful, there are still aspects of explainability that could be further explored using traditional XAI methods. For example, it may be of interest to forensic analysts using the system which locus, or perhaps which peaks were the most influential in assigning a particular NoC over another. Shapley values [34] could be explored for this task. Shapley values were first developed in cooperative game theory to assess the contribution of each player to a game. They have since become a way of assessing the contribution of each input to an assignment within a machine learning context (for a NoC assignment example see [18]). Shapley values are $2^N$ hard problems, so as the number of inputs increases the practicality of exhaustively searching the sample space becomes impractical. Even at a per-locus basis it is likely that 24 inputs would be beyond the practical application of a standard Shapley value calculation. Recent work on the application of Shapley value calculations to classification of electrophoretic data in forensic science has been developed [19] which utilises a focussing approach. This starts by examining large 'super-pixels' of the problem and then focussing down on just those that show some contribution to the assignment and breaking them into smaller and smaller sections in order to achieve the desired resolution. That same process may work for the deepNoC application by starting on a dye-by-dye basis (of which there are six, a value easily handled computationally), and then focussing on the loci only within the dyes that show some contribution to the NoC assignment.

4.2.3 – Sensitivity analysis

Given the simulation pipeline used in the development of deepNoC, it is likely that the pre-fine-tuned model is specific to the profile reading ANN used, the performance of the laboratory in generating DNA profiles, and the ethic population of the individuals in the mixture. We have shown how fine-tune training can overcome these differences when we carried out training on the ProvedIt dataset. Specifically;

- the ProvedIt profiles were produced in a laboratory that was different to the laboratory on which the simDNAmixtures profile simulation tool was calibrated
- The profiles were read in FaSTR DNA using peak detection algorithms that different from those used in profile simulation
- The ProvedIt profiles were constructed from individual's DNA from the US, whereas the simulation was sampling profiles from a population database of individuals from Australia

Despite these differences the fine-tune training was able to quickly update deepNoC from minimal fine-tuning samples. This suggests that effects of the listed points may have only a minor effect. However, in casework the ethnicity of the donors to a DNA profile are unknown and may occur regions with multiple genetically diverse populations all living in the same area. It is unknown how much of an effect training deepNoC on one population and applying it to another would have on NOC assignment accuracy. The sensitivity of classification to population misalignment is something that needs to be investigated before the system could be put into use.

A second type of sensitivity that could be explored is the sensitivity of the assignment to small changes in the DNA profile peak heights. The stochastic nature of peak heights is a well-known phenomenon in forensic biology [3], with each replicate PCR, or replicate electrophoretic run, or separate injection on the electrophoretic run having an effect [35]. If deepNoC is performing robustly then it would be expected that small changes in peak heights (or even changes in the presence or absence of minor peaks) should have a limited effect on NoC assignment. This could be tested by simulating small stochastic effects on the peaks within a profile being classified and noting the change in classification probabilities.

4.2.4 – A question of user requirements

Several authors within forensic biology have commented on the concept of 'apparent NoC' [21, 36], which describes the concept that there may be contributors to a DNA that are present at such minor levels, or in such a way, that they do not affect the fluorescent signal from other contributors. An artificial (but conceptually useful) example would be to consider a DNA sample that is constructed in a laboratory from two contributions, one with DNA in 100-fold excess of the other. If only a small amount of this sample was used to generate a DNA profile, all that would be detected is the DNA from the main donor. The 'target NoC' (i.e. as experimentally designed) is two, but the apparent NoC is one. Arguably the NoC which is most appropriate to analyse the profile as is the apparent NoC.

This raises an interesting point, as there will be profiles for which it is clear that the target NoC is desired and there are profiles where it seems clear that the apparent NoC is desired, but there is a large grey zone during which humans will lose the ability to 'see' the effects of a contributor, before they become truly non-existent in the data. This is where systems such as deepNoC are most useful, although there will be a required leap of faith in the system on the part of the analyst to trust that it has identified some information that they themselves cannot. It is clear from the results in Table 3 that deepNoC (and other systems) outperform the ability of people to assign NoC, and so this faith would seem to be well supported. However, conveying this faith in ability of an algorithm to a court, or a vigorous defence challenge, may be difficult. It will require both explainability (as has been built into deepNoC's architecture) and careful validation of sensitivities (as explained earlier) and performance.

**5.0 - Conclusion**

A deep learning neural network system, deepNoC, was developed to assign the number of contributors to an STR DNA profile. The development of deepNoC overcame limitations with previous systems, specifically that of limited data, by utilising a profile simulation pipeline to generate a large number of profiles to train on. The system was then able to be fine-tuned using a much smaller number of laboratory-generated profiles. The deepNoC system outperformed any equivalent system of assigning a number of contributors, with an overall accuracy of approximately 90% for 1 to 5 contributors and 72% for 1 to 10 contributors. The deepNoC system was also designed with explainability in mind, so that the multiple outputs of the system can be graphically displayed in order to show how different parts of the profile were viewed within the overall classification.

There are several avenues in which the deepNoC system can be expanded or investigated, however the initial results show a high performance that would allow the development of automated profile analysis pipelines in forensic facilities.

**Supplementary material**

1 – detailed diagram of deepNoC architecture.

**Acknowledgements**

Points of view in this document are those of the author and do not necessarily represent the official position or policies of their organisations. This work was supported by a grant awarded from the Australian New Zealand Policing Advisory Agency National Institute of Forensic Sciences research funding support scheme.